\documentclass[graybox, envcountchap]{svmult}

\usepackage{mathptmx}        % selects Times Roman as basic font
\usepackage{helvet}          % selects Helvetica as sans-serif font
\usepackage{courier}         % selects Courier as typewriter font
\usepackage{makeidx}         % allows index generation
\usepackage{graphicx}        % standard LaTeX graphics tool
\usepackage{multicol}        % used for the two-column index
\usepackage[bottom]{footmisc}% places footnotes at page bottom

\makeindex             % used for the subject index
\begin{document}

\frontmatter%%%%%%%%%%%%%%%%%%%%%%%%%%%%%%%%%%%%%%%%%%%%%%%%%%%%%%

%%%%%%%%%%%%%%%%%%%%%%% dedic.tex %%%%%%%%%%%%%%%%%%%%%%%%%%
%
% sample dedication
%
% Use this file as a template for your own input.
%
%%%%%%%%%%%%%%%%%%%%%%%% Springer %%%%%%%%%%%%%%%%%%%%%%%%%%

\begin{dedication}
Use the template \emph{dedic.tex} together with the Springer document class SVMono for monograph-type books or SVMult for contributed volumes to style a quotation or a dedication\index{dedication} at the very beginning of your book in the Springer layout
\end{dedication}

%%%%%%%%%%%%%%%%%%%%%%foreword.tex%%%%%%%%%%%%%%%%%%%%%%%%%%%
% sample foreword
%
% Use this file as a template for your own input.
%
%%%%%%%%%%%%%%%%%%%%%%%% Springer %%%%%%%%%%%%%%%%%%%%%%%%%%

\foreword

Use the template \textit{foreword.tex} together with the Springer document class SVMono (monograph-type books) or SVMult (edited books) to style your foreword\index{foreword} in the Springer layout. 

The foreword covers introductory remarks preceding the text of a book that are written by a \textit{person other than the author or editor} of the book. If applicable, the foreword precedes the preface which is written by the author or editor of the book.

\vspace{\baselineskip}
\begin{flushright}\noindent
Place, month year\hfill {\it Firstname  Surname}\\
\end{flushright}

%%%%%%%%%%%%%%%%%%%%%%preface.tex%%%%%%%%%%%%%%%%%%%%%%%%%%%%%%%%%%%%%%%%%
% sample preface
%
% Use this file as a template for your own input.
%
%%%%%%%%%%%%%%%%%%%%%%%% Springer %%%%%%%%%%%%%%%%%%%%%%%%%%

\preface

Use the template \emph{preface.tex} together with the Springer document class SVMono (monograph-type books) or SVMult (edited books) to style your preface in the Springer layout.

A preface\index{preface} is a book's preliminary statement, usually written by the \textit{author or editor} of a work, which states its origin, scope, purpose, plan, and intended audience, and which sometimes includes afterthoughts and acknowledgments of assistance. 

When written by a person other than the author, it is called a foreword. The preface or foreword is distinct from the introduction, which deals with the subject of the work.

Customarily \textit{acknowledgments} are included as last part of the preface.

\vspace{\baselineskip}
\begin{flushright}\noindent
Place(s),\hfill {\it Firstname  Surname}\\
month year\hfill {\it Firstname  Surname}\\
\end{flushright}

%%%%%%%%%%%%%%%%%%%%%%acknow.tex%%%%%%%%%%%%%%%%%%%%%%%%%%%%%%%%%%%%%%%%%
% sample acknowledgement chapter
%
% Use this file as a template for your own input.
%
%%%%%%%%%%%%%%%%%%%%%%%% Springer %%%%%%%%%%%%%%%%%%%%%%%%%%

\extrachap{Acknowledgements}

Use the template \emph{acknow.tex} together with the Springer document class SVMono (monograph-type books) or SVMult (edited books) if you prefer to set your acknowledgement section as a separate chapter instead of including it as last part of your preface.

\tableofcontents
%%%%%%%%%%%%%%%%%%%%clist.tex %%%%%%%%%%%%%%%%%%%%%%%%
%                                                    
% sample list of contributors and their addresses    
%                                                    
% Use this file as a template for your own input.    
%                                                    
%%%%%%%%%%%%%%%%%%%%%%%% Springer %%%%%%%%%%%%%%%%%%%%
\contributors

\begin{thecontriblist}
Firstname Surname
\at ABC Institute, 123 Prime Street, Daisy Town, NA 01234, USA, \email{smith@smith.edu}
\and
Firstname Surname
\at XYZ Institute, Technical University, Albert-Schweitzer-Str. 34, 1000 Berlin, Germany, \email{meier@tu.edu}
\end{thecontriblist}
%%%%%%%%%%%%%%%%%%%%%%acronym.tex%%%%%%%%%%%%%%%%%%%%%%%%%%%%%%%%%%%%%%%%%
% sample list of acronyms
%
% Use this file as a template for your own input.
%
%%%%%%%%%%%%%%%%%%%%%%%% Springer %%%%%%%%%%%%%%%%%%%%%%%%%%

\extrachap{Acronyms}

Use the template \emph{acronym.tex} together with the Springer document class SVMono (monograph-type books) or SVMult (edited books) to style your list(s) of abbreviations or symbols in the Springer layout.

Lists of abbreviations\index{acronyms, list of}, symbols\index{symbols, list of} and the like are easily formatted with the help of the Springer-enhanced \verb|description| environment.

\begin{description}[CABR]
\item[ABC]{Spelled-out abbreviation and definition}
\item[BABI]{Spelled-out abbreviation and definition}
\item[CABR]{Spelled-out abbreviation and definition}
\end{description}

\mainmatter%%%%%%%%%%%%%%%%%%%%%%%%%%%%%%%%%%%%%%%%%%%%%%%%%%%%%%%
%%%%%%%%%%%%%%%%%%%%%part.tex%%%%%%%%%%%%%%%%%%%%%%%%%%%%%%%%%%
% 
% sample part title
%
% Use this file as a template for your own input.
%
%%%%%%%%%%%%%%%%%%%%%%%% Springer %%%%%%%%%%%%%%%%%%%%%%%%%%

\begin{partbacktext}
\part{Part Title}
\noindent Use the template \emph{part.tex} together with the Springer document class SVMono (monograph-type books) or SVMult (edited books) to style your part title page and, if desired, a short introductory text (maximum one page) on its verso page in the Springer layout.

\end{partbacktext}

\title*{Language-agnostic Code-Switching in Sequence-To-Sequence Speech Recognition}
\author{Enes Yavuz Ugan$^1$,Christian Huber$^1$, Juan Hussain $^1$ and Alexander Waibel$^{1,2}$}
\institute{$^1$Interactive Systems Lab, Karlsruhe Institute of Technology, Karlsruhe, Germany\\
  $^2$Carnegie Mellon University, Pittsburgh PA, USA\\
  firstname.lastname@kit.edu, alexander.waibel@cmu.edu}
%\institute{Name of First Author \at Name, Address of Institute, \email{name@email.address}
%\and Name of Second Author \at Name, Address of Institute \email{name@email.address}}
\maketitle

\abstract*{
Code-Switching (CS) is referred to the phenomenon of alternately using words and phrases from different languages. 
While today's neural end-to-end (E2E) models deliver state-of-the-art performances on the task of automatic speech recognition (ASR) it is commonly known that these systems are very data-intensive. 
However, there is only a few transcribed and aligned CS speech available. 
To overcome this problem and train multilingual systems which can transcribe CS speech, we propose a simple yet effective data augmentation in which audio and corresponding labels of different source languages are concatenated.
By using this training data, our E2E model improves on transcribing CS speech.
It also surpasses monolingual models on monolingual tests.
The results show that this augmentation technique can even improve the model's performance on inter-sentential language switches not seen during training by 5,03\% WER.
}
\abstract{
Code-Switching (CS) is referred to the phenomenon of alternately using words and phrases from different languages. 
While today's neural end-to-end (E2E) models deliver state-of-the-art performances on the task of automatic speech recognition (ASR) it is commonly known that these systems are very data-intensive. 
However, there is only a few transcribed and aligned CS speech available. 
To overcome this problem and train multilingual systems which can transcribe CS speech, we propose a simple yet effective data augmentation in which audio and corresponding labels of different source languages are concatenated.
By using this training data, our E2E model improves on transcribing CS speech.
It also surpasses monolingual models on monolingual tests.
The results show that this augmentation technique can even improve the model's performance on inter-sentential language switches not seen during training by 5,03\% WER.
}
%\abstract{Each chapter should be preceded by an abstract (10--15 lines long) that summarizes the content. The abstract will appear \textit{online} at \url{www.SpringerLink.com} and be available with unrestricted access. This allows unregistered users to read the abstract as a teaser for the complete chapter. As a general rule the abstracts will not appear in the printed version of your book unless it is the style of your particular book or that of the series to which your book belongs.\newline\indent
%Please use the 'starred' version of the new Springer \texttt{abstract} command for typesetting the text of the online abstracts (cf. source file of this chapter template \texttt{abstract}) and include them with the source files of your manuscript. Use the plain \texttt{abstract} command if the abstract is also to appear in the printed version of the book.}

\section{Introduction}
\label{sec:introduction}
Due to increasing globalization, a growing number of people move to foreign countries to make a living an example would be Germany which shows an increase from 9.107.895 foreign population in 2015 to 11.817.790 in 2021 \cite{destatis}.
As these people start learning a new language, this can result in Code-Switching (CS), which is referred to as the change between languages while speaking. % \cite{mesthrie2009introducing}.
An example of German-English CS would be the phrase 'Das war sehr strange' ('That was very strange').

From a linguistic perspective, CS can be divided into multiple categories \cite{poplack1980sometimes}:
\begin{itemize}
    \item Inter-sentential CS: The switch between languages happens at sentence boundaries.
    \item Intra-sentential CS: Here the second language is included in the middle of the sentence.
        Additionally, the word borrowed from the second language can happen to be adapted to the grammar of the matrix language as well.
    \item Extra-sentential CS: In this case, a tag element from a second language is included, for example at the end of a sentence. This word is more excluded from the main language.
\end{itemize}
As these developments can result in growing numbers of multilingual communities and individuals the need for dialogue and ASR systems capable of processing such CS data is very important.
Despite occurring frequently, CS poses a great challenge for all neural-network-based end-to-end ASR models.
As of today, there are only a few CS data available for a very limited number of languages. 
Some example corpora available are \cite{amazouz2017addressing} for CS between French and Algerian speech, \cite{lyu2010analysis} containing utterances switching between Mandarin and English, and \cite{chan2005development} having gathered data with CS between English and Cantonese.

As an exemplary case, in this work, we focus on developing a multilingual ASR system capable of transcribing CS utterances between German and English. 
%This is a prevalent scenario considering the increased amount of Arabic-speaking people in Germany.
Considering the increased amount of Arabic-speaking people in Germany another common language that is mixed with German is Arabic and thus we included it as a third language.
These languages are specifically interesting to analyze as German and English are from the same Indo-European language family while Arabic is part of the Afro-Asiatic language family.
As training data is not available in our scenario, we conduct multiple experiments using a straightforward CS data augmentation technique.
Specifically, we present the following contributions:

First, we present a simple yet effective data augmentation technique designed for CS models in data-scarce scenarios, by simply concatenating multilingual sources and corresponding targets without any language information. 
Second, we perform an extensive evaluation of our model on intra- \& inter-sentential CS test sets, as well as monolingual ones. 
Next to using publicly available test sets, we conduct our evaluation on artificially generated as well as our in-house collected data.
Our experiments yield interesting results section \ref{sec:experiments-results}, including 1) enabling the Sequence-to-Sequence (S2S) model to reliably transcribe CS utterances, 2) improving the performance of the multilingual model on monolingual test sets and 3) the capability of transcribing CS utterances for language pairs not switched during training.

\section{Related Work}
\label{sec:related-work}
As transcribing Code-Switching utterances inherently needs an ASR model which is multilingual to some degree, we want to refer to some of the early work in this research area such as \cite{stuker2003multilingual}, \cite{schultz2001experiments}, \cite{schultz2001language}, \cite{mussakhojayeva2021study}, \cite{watanabe2017language} and \cite{muller2017language}.

As there are only a few CS data available there has not been too much research for many language pairs especially if there is no data present. 
Some of the language pairs addressed are Frisian-Dutch \cite{yilmaz2016investigating}, Malay-English \cite{ahmed2012automatic}, dialectal Arabic-English \cite{hamed2022investigations}, different Indian languages with English in \cite{diwan2021multilingual}, Korean-English \cite{lee2021phonetic}, Japanese-English \cite{nakayama2018speech},\cite{nakayama2019zero}, and Mandarin-English \cite{weiner2012integration}, \cite{luo2018towards}, \cite{zeng2018end}, \cite{chang2018code}.

Most of the work on Code-Switching focuses on language pairs with some available CS data.
In \cite{shan2019investigating} the authors aim at solving the problem of code-switching using a multi-task learning (MTL) approach.
The authors investigate training a model predicting a sequence of labels as well as predicting language identifiers at different levels of the architecture. 
They also report that first training with monolingual data and fine-tuning it with CS speech improves their performance. 
In \cite{shah2020learning} the writers analyzed the effect of fine-tuning toward CS data on monolingual ASR. 
They show that fine-tuning a model on CS and monolingual data yields a better overall Word Error Rate (WER)s than when only using Code-Switching data. 
In \cite{li2019towards} the authors propose to train two separate models. 
One CTC model for speech recognition and another one for frame-level language prediction. 
During decoding, if the current frame has a very high probability for the blank symbol the blank label is emitted, otherwise the output probabilities of English tokens are multiplied by the probability of this frame being English and the Chinese labels are multiplied by the probability of this frame being Chinese speech. 
While improving the CS WER they report a decrease in monolingual speech recognition performance.
To improve the model's performance on CS speech in \cite{zhang2022reducing} the authors propose language-related attention mechanisms to profit more from using monolingual data, next to CS ones.

Other works try utilizing CS training data in order to use it for data augmentation. 
In that way, they aim at improving the performance by utilizing more than the original available CS data.
%In \cite{seki2018end} the authors train a hybrid attention/connectionst temporal classification (CTC) network which first classifies which language is going to be transcribed followed by the transcription itself. 
In \cite{yilmaz2018acoustic} the authors used a separate TDNN-LSTM \cite{peddinti2017low} as an acoustic model, as well as a separate language model. 
Thus they were able to utilize CS speech-only data for enhancing the acoustic model.
They also enhanced their language model separately using artificially created text-only CS data.
Thus they were able to improve over a baseline model only trained with the original CS data.
%Using different approaches, they also artificially created text-only CS data, for enhancing their language model.
Another work \cite{guo2018study}, is also using a semi-supervised approach focusing on improving the lexicon and the acoustic model of an HMM-based ASR model. First, they extend the lexicon to realize no out-of-vocabulary for their training data. As in their CS data, English words have accented pronunciations, afterwards, they use a phonetic level decoding to learn adapted pronunciations of words. Additionally, they used audio of transcriptions with high Word Matched Error Rate in order to improve their acoustic model in a semi-supervised fashion. Each of their steps yields improvements in the English-Chinese CS setup on which they evaluated.
In \cite{du2021data} the authors propose three different data augmentation algorithms. 
They apply audio splicing, meaning they randomly insert audio segments of the same speaker in a different language into the original utterance. 
The other two approaches are randomly inserting or translating a word in the source text and generating the corresponding audio using a TTS system.
Here, the TTS system is trained with CS data, and also the word alignments needed for audio splicing were retrieved using an HMM-GMM ASR system trained on the initial CS data.

However, some of the earlier work also considered developing ASR models without the use of transcribed CS data.
In \cite{seki2018end} the authors train a hybrid attention/connectionst temporal classification (CTC) network which first classifies which language is going to be transcribed followed by the transcription itself. 
%Other approaches try utilizing synthetically generated CS speech. 
In \cite{nakayama2019zero} the authors address the task of CS in ASR and TTS using a semi-supervised learning approach using the machine speech chain. 
In their approach first, an ASR and a TTS model are trained separately using monolingual data.
Afterward, they utilize speech-only data by first transcribing it and re-synthesize the transcription in order to update the TTS model.
CS text-only data is utilized by synthesizing the transcript and then transcribing it afterward. 
That way the ASR model gets trained for the CS task.
The authors also use speaker embeddings in order to make sure the synthesized speech is the same speaker as is in the input.
Their strategy improves CS performance without using any transcribed CS data but improves even further if some paired CS data is used as well.
%The ASR model is trained using CS text. 
%First the CS text is synthesized using a TTS model which receives the text and character level language information as well as some random speaker vector. 
%The synthesized speech is than transcribed and the ASR model weights are updated. 
%Given CS speech only data the ASR model generates a sequence of characters as well as language tokens, additionally a speaker vector is extracted separately. 
%This three values are passed to the TTS system and according to the error resulting in the difference of the input and output speech the TTS model gets updated. 
%They report relative improvements of 0,4\%, 1,45\%, 1,02\% WER if there no paired CS data is available. 
%They also report relative improvements in the zero-shot scenario  0,31\%, 1,3\%, 0,89\% for inter-sentential CS pairs of English-Japanese, Japanese-Chinese and Chinese-English. 
%\TODO{I can put improvements in evaluation as comparison.}
Another interesting data-augmentation approach is presented in \cite{hussein2022code}. 
They propose an approach consisting of multiple steps, in order to generate artificial CS text-only data.
First, some Arabic text is translated into English.
The Arabic script is morphologically segmented in order to calculate a better alignment between the translations.
Afterward, a sentence-level constituent parse tree is generated and the CS data is generated according to the Equivalence Constraint theory described in \cite{pratapa2018language}.
This data is used to improve the Language Model of their HMM-GMM ASR model.

As can be read, most of the previous work considers cases in which some CS training data is available. 
We were interested in training a S2S model, without any real CS data, which is the more common case considering the data available.
Our model should not be explicitly trained to predict languages but should do so implicitly by predicting the right labels, which in our case are Byte pair encoding tokens.
Additionally, we analyzed the effects of different data augmentation constraints on the model's performance.

\section{MODEL}
\label{sec:model}
\begin{figure}[b]
\sidecaption
\includegraphics[scale=0.22,page=2,trim={18cm 3cm 23cm 0cm},clip]{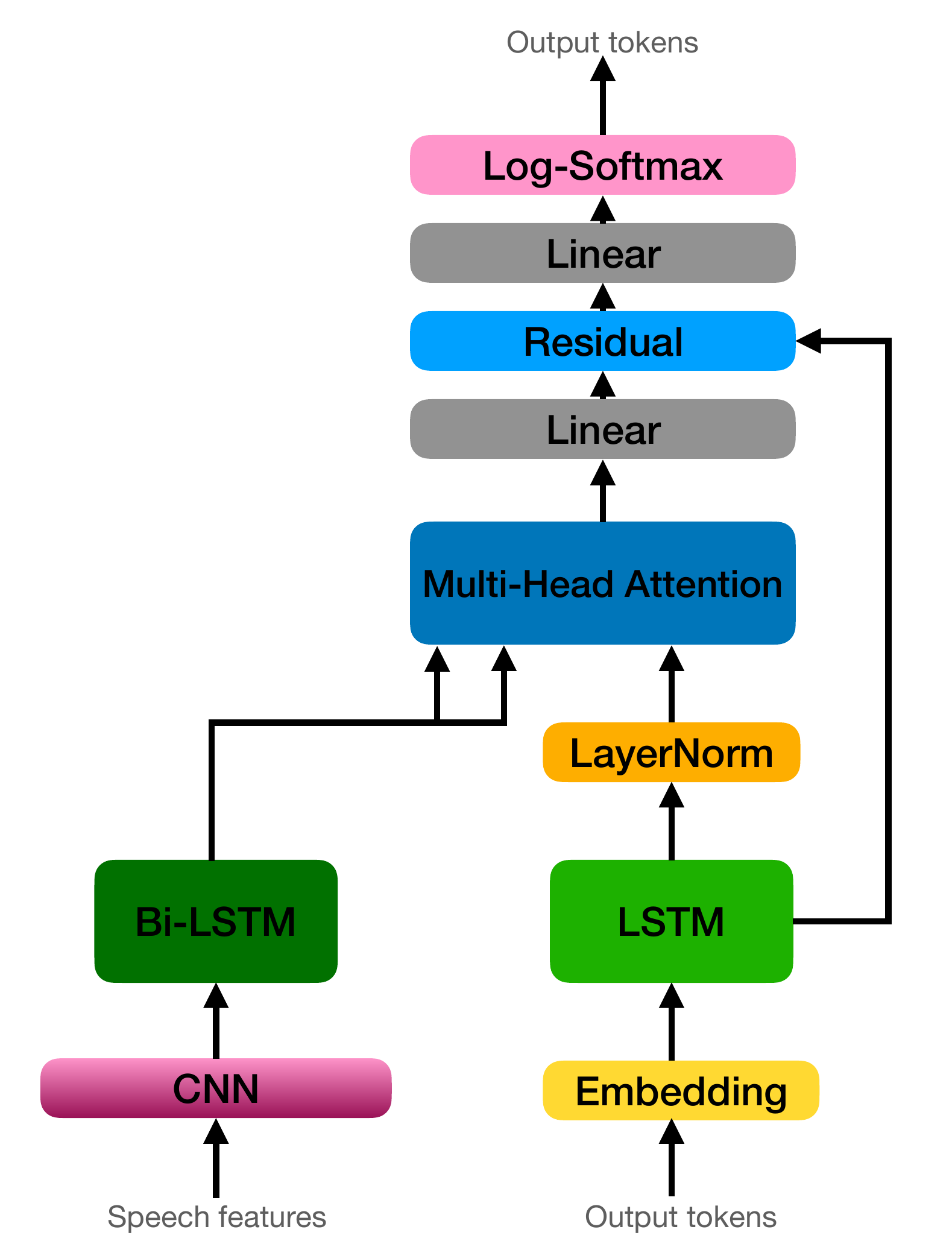}
\caption{Abstract architecture of the encoder-decoder-based Sequence-to-Sequence Model used in our experiments.}
\label{fig:LSTM-Model}       % Give a unique label
\end{figure}

For our experiments, we used a S2S encoder-decoder-based model, as described in \cite{nguyen2020improving}. 
An abstract description of the neural network would look like this:
\begin{eqnarray*}
enc =& \textit{Bi-LSTM}(\textit{CNN}(log\-Mel\_Spectrum)) \\
tgt\_emb =& \textit{LSTM}(\textit{Embedding}(out\_tokens))\\
dec =& (\textit{MHA}(enc,enc,tgt\_emb)+tgt\_emb)\\
output =& \textit{log\_softmax}(dec)
\end{eqnarray*}
In more detail, the model consists of a two-layer Convolutional Neural Network (CNN) applying 32 channels. 
We choose the window size of three over the frequency, as well as the time domain.
A stride of two was applied resulting in a spectrogram down-sampled by a factor of four. 
After down-sampling a six-layer bidirectional LSTM is adopted.
The decoder consists of an embedding layer followed by a two-layer unidirectional LSTM.
The hidden size for all LSTMs was set to 1024.
The output of the encoder and decoder LSTMs are used to calculate a context vector using a multi-head cross-attention mechanism \cite{vaswani2017attention} with eight heads.
After applying a residual connection with the decoder LSTM output, a projection layer is used to project the hidden dimension size to the size of the vocabulary. 
The architecture is depicted in Fig.~\ref{fig:LSTM-Model}.

As input, we use 40-dimensional log-Mel features calculated on frames of 25 ms with a stride of 15 ms. 
In contrast to some of the previous other works, we use one Byte pair encoding (BPE) \cite{gage1994new} calculated on all three languages. 
This means we have in total 4000 labels for all languages. 
When calculating the BPE we made sure to use the same amount of text data for the three languages. 
The resulting BPE contains 2553 English, German, and 1444 Arabic tokens.
The other three tokens are the unknown, start of sequence, and end of sequence tokens.
The labels for the monolingual experiments were calculated on monolingual text data.
We decided to use BPE tokens as they have shown to yield good results in the ASR task.
Another reason we did not choose some common representation space for the Latin alphabet and the Arabic abjad is that we aim at printing the transcription in the correct language without any additional systems needed.
If we transliterated Arabic into Latin script the question would arise if the hypothesis is actually an Arabic or English/German transcription of the speech.

The same number of model parameters are used in all our experiments.
1024 dimensional LSTMs are trained using Adam optimizer \cite{kingma2014adam} with a maximum learning rate of 0,002 and 8000 warm-up steps. 
After each epoch, the perplexity is used to determine if the model improved or decreased in performance.
The validation performance was determined by adding the perplexity of the monolingual validation sets of each language, as well as a pseudo-CS validation set generated using the same three validation sets by applying the algorithm explained in section \ref{subsec:data-augmentation}.
An early abortion was applied if there were no improvements over five epochs. 
For tests, the epoch with the lowest validation perplexity during training was chosen.

\section{ DATA }
\label{sec:data}
As already mentioned in Section \ref{sec:introduction} we used three languages in this work, namely Arabic, German and English. 
The English training data is made up of How2 \cite{sanabria2018how2} and TED-LIUM (TED) \cite{rousseau2012ted} data sets. 
For the German training data we used  Common Voice (CV) \cite{ardila2019common}, Europarl \cite{koehn2005europarl}, Lect. a data set of recorded lectures and interviews%by KIT%
, and  Mini-international Neuropsychiatric Interview (MINI)-Data. 
As Arabic training data, we used MGB2 (Alj.) data from \cite{ali2016mgb} and MINI data \cite{huber2020supervised}. 
An overview of our training data is given in Table~\ref{tab:tr-data}.

\begin{table}%[t!]
\caption{ Data used during training.}
\label{tab:tr-data}
%\small
\begin{tabular}{p{2cm}p{2.4cm}p{2cm}p{4.9cm}} 
\hline\noalign{\smallskip}
 Language & Corpus & Speech [h] & Utterances\\ 
 \noalign{\smallskip}\svhline\noalign{\smallskip}
 English (EN)   & How2 & 345 & 210k \\ 
                        & TED & 439 & 259k \\

 German (DE) & CV & 314 & 196k \\
                     & Europarl & 46 & 20k \\
                     & Lect. & 504 & 353k \\
                     & MINI-Data & 1  & 498 \\

 Arabic (AR) & Alj. & 1127 & 375k \\
                     & MINI-Data & 39 & 9k \\
\hline
\end{tabular}
\end{table}
For our tests, we have monolingual test sets for each language. 
The Alj.2h data was dialect-free Arabic data extracted as explained in \cite{hussain-etal-2020-german}. 
In order to evaluate the CS performance, we generated a test set (artificial) by applying the data augmentation technique, using CV, Alj.2h, and WSJ test sets, as described in Section \ref{subsec:data-augmentation}. 
For intra-sentential CS with German as the matrix and English as the embedded language, we use our in-house Lect. test set where English words have been manually annotated.
%number of <ENG> per word: 439/14741 -> 0.029780883250797096(2,98%); number of <ENG> per line: 439/204 -> 2.1519607843137254
We combined this data with a small set of read speech, collected by us. 
%for 25 out of 89 utts: ~2,88 En per utt; 12,4 words per utt; (0.23225806451) 23,22 % English words
This is depicted as Deng. in Table~\ref{tab:test-data}.
Here the overwhelming amounts are German words and only 4,5\% are English.
We also tested our models on the German-English intra-sentential CS test set derived from the Spoken Wikipedia Corpus (SWC) provided by \cite{khosravani2021evaluation}, depicted as SWC-CS. 
Detailed information about our test sets can be taken from Table~\ref{tab:test-data}. 
Both of these intra-sentential sets have German as the matrix language with English words embedded. 
tst-inter is an inter-sentential CS set we derived from MuST-C (tst-COMMON) \cite{di2019must} data. 
At sentence boundaries, the sentence was continued in either English or German in a CS manner.
These sentences were then read by two persons. 
We also collected a German-English CS test set (D-E-CS) and a German-Arabic test set (D-A-CS) which contain switches at dependent and independent clauses and as such contain longer intra-sentential CS data, as well as inter-sentential Data.
This data was generated by using our Lect. test set and tst-common and translating clauses into the respective language.
Afterward, the utterances were read by 4 and 2 speakers for the D-E-CS and D-A-CS respectively, using the TEQST tool \footnote{https://github.com/teqst}.
In D-E-CS 57,3\% of the clauses are German and 42,7\% are English.
In D-A-CS 50\% of the clauses are German and 50\% are Arabic.
While the speakers reading in the D-E-CS set were of German origin, the speakers reading D-A-CS originated from Arabic countries.
Participants reading utterances containing English were L1 in German and L2/B1 in English.
Speakers recording text with Arabic as a language pair were L1 in Arabic and L3/B1 in German.
\begin{table}
\caption{Data used for testing.}
\label{tab:test-data} 
\begin{tabular}{p{2cm}p{2.4cm}p{2cm}p{4.9cm}} 
\hline\noalign{\smallskip}
 Language & Corpus & Speech [h] & Utterances\\ 
  \noalign{\smallskip}\svhline\noalign{\smallskip}
    English  & TED & 3 & 1k\\ 
           %             & WSJ & 1 & 503 \\
   
    German & CV & 25 & 15k \\
                     & Lect. & 5,2 & 5k \\

    Arabic & Alj. & 10 & 5k \\
                     & Alj.2h & 2 & 1k \\

    Intra-sent. %%& Lecture-CS & 5,6h & 6432 \\
                                & SWC-CS \cite{khosravani2021evaluation} & 34,1 & 12437 \\
                                & Deng. & 0,95 & 293\\ %& 6min & 89 Helens; only-cs Lect.: 1,9 & 204\\

    Inter-sent. & artificial & 1,9 & 1687 \\
                            & tst-inter & 0,85 & 284 \\

    Mix-sent. & D-E-CS & 1,42 & 562 \\
                          & D-A-CS & 1,09 & 398 \\

\hline
\end{tabular}
\end{table}

\section{ APPROACH }
\label{subsec:data-augmentation}
Inspired by \cite{seki2018end}, we applied a concatenation technique to generate our CS data. 
We concatenate the log-Mel features of different languages after each other. 
For the target labels, we also concatenate the respective labels after each other.
We want to note, that the speakers are not the same in each language, as such the resulting data can not be considered CS data but more like pseudo-CS data.
As it turned out to be an important factor in training the model, we enable to set a specified relative amount of CS utterances in the training set.
During data augmentation, we only have two restrictions. 
First, we limit the amount of CS data to a specific percentage. 
As we need to define a restriction on how long concatenated utterances are allowed to be, we analyzed our monolingual training data.
We saw that most utterances are less than 10 seconds long. 
Thus, our second restriction is that we limit the length of the CS data. 
25\% of the CS data was made to be five seconds, another 25\% up to 10 seconds another 25\% 15 seconds long. 
Utterances of 20 and 25 seconds each made up 12,5\% of the newly generated CS data. 
For each of the above-mentioned time ranges, we generate CS data the following way.
First, a language is chosen randomly with an equally distributed probability. 
Afterward, an utterance is randomly picked out of all the sequences in that language. 
These steps are repeated until the CS duration of the sequence is up to two seconds shorter than the current time range. 
As we can see our data augmentation does not add any information about the language being transcribed and has no major restrictions. 
Keeping the process simple benefits an easier and more general usage of this approach. 
As, in intra-sentential CS cases, a word of the embedded language can be adapted to the grammar of the matrix language we believe that not predicting the language explicitly during decoding is also beneficial for training the model in a more general way. 

An example input feature is shown in the following Fig.~\ref{fig:PCS}. 
In a) a monolingual German utterance which was randomly picked as described above is depicted. 
In b) a second utterance, this time a monolingual English one was selected. 
The algorithm we propose now concatenates their logarithmic Mel features as well as their transcript and passes them as the model input and the ground truth for teacher forcing.

\begin{figure}[b]
%\sidecaption
\includegraphics[scale=0.18,page=1,trim={0cm 2cm 5cm 5cm},clip]{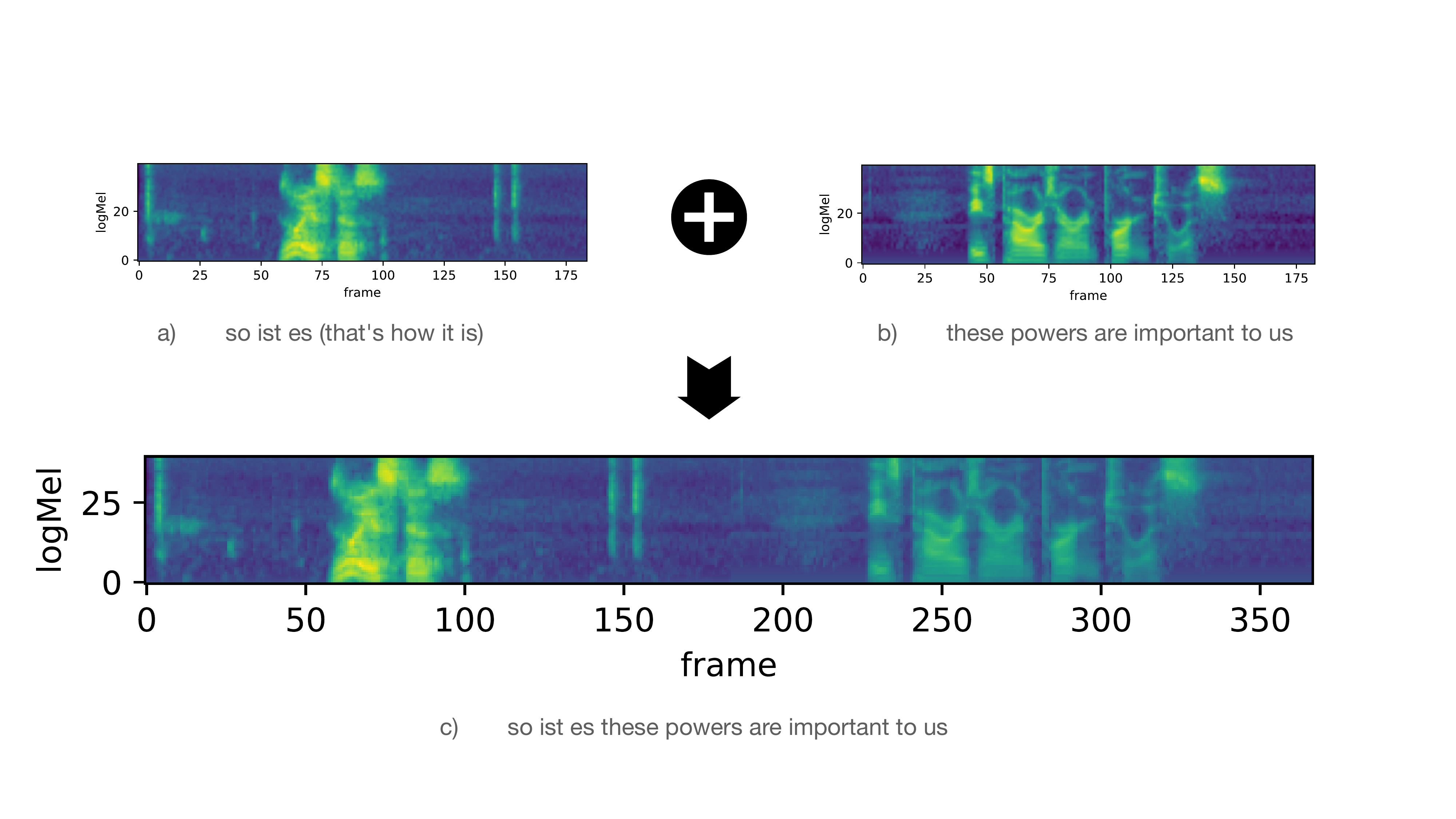}
\caption{a) Monolingual German utterance. b) Monolingual English utterance. c) The concatenated features resulting in our pseudo-CS data for the target transcript "so ist es these powers are important to us". }
\label{fig:PCS}       % Give a unique label
\end{figure}

%\begin{figure}[b]
%%\sidecaption
%\includegraphics[scale=.35]{templates/images/f1mel40Test.pdf}
%\includegraphics[scale=.35]{templates/images/s1mel40Test.pdf}
%
%\caption{logMel features of the audio "so ist es" on the left hand side, and "these powers are important to us" on the right hand side.}
%\label{fig:deenMel}       % Give a unique label
%\end{figure}
%
%\begin{figure}[b]
%%\sidecaption
%\includegraphics[scale=.65]{templates/images/mel40Test.pdf}
%\caption{The concatenated logMel features resulting in our pseudo CS data for the transcript "so ist es these powers are important to us". }
%\label{fig:PCS}       % Give a unique label
%\end{figure}

\section{ EXPERIMENTS }
\label{sec:experiments-results}

\subsection{ Baselines }
\label{subsec:baseline}
As for baselines, we trained four different models.
One monolingual model for each of the languages Arabic (Mono-Ar), German (Mono-De), and English (Mono-En). 
The fourth baseline is a multilingual model (Mult.), which we trained using the concatenated data set of the three languages. 
In Table~\ref{tab:res-bas-mono} the WER performances of these models are reported on monolingual test sets.
%As expected each monolingual model is only able to transcribe data in its respective language. 
While the multilingual model can transcribe all languages, a drop in performance can be seen in all tests when compared with the monolingual counterpart.

The performance of our baseline models on multilingual CS data is provided in Table~\ref{tab:res-bas-cs}.
For Mono-Ar and Mono-En, it can be seen that the performance on CS data is quite bad. 
On our intra-sentential tests, the monolingual German model has the best results, this is due to German being the matrix language and as mentioned in \ref{sec:data}, English words are only embedded sporadically in these utterances.
%Looking at the inter-sentential and the mixed CS test set (D-E-CS), it is visible that Mono-De and Mono-En perform very badly as well. 
Looking at tst-inter and the D-E-CS, it is visible that Mono-De and Mono-En perform very poorly as well. 
Mono-De has a slightly lower WER probably because there are more German clauses in the test set than English ones.
%Another reason is, that it was able to decode some of the English words as well, which hints that there are some English words in our BPE calculation and training data.
%Another reason is, that it was able to decode some of the English words as well, which hints that there are some English words in our training data.
%the strong baseline performance of the Mono-De model on Denglish data hints, that nowadays it is so common to embed English words in German speech that our German data set contains such appearances, although they are not marked as such.

While the multilingual model decreased the performance by relative 7,69\% WER on the intra-sentential CS, it was able to outperform the Mono-De model by relative 52,58\% WER on tst-inter and relative 43,67\% WER on D-E-CS.

Interestingly on D-A-CS, we can see similar scores for Mono-DE and Mult.. 
Looking at the transcripts we see that the multilingual model only transcribes parts of the utterance in one of the two languages. 
Similar to Mono-DE which only transcribes German parts of the utterance. 
Compared to the improvements in D-E-CS this shows that sharing the language scripts can have major benefits for multilingual models.

\setlength{\tabcolsep}{2.4pt}
%\begin{table}[htbp]
\begin{table}
\caption{Results of baseline models on monolingual test sets. Results are reported in WER\%.}
\label{tab:res-bas-mono}

\small
\begin{tabular}{p{1.7cm}p{1.8cm}p{1.8cm}p{1.8cm}p{1.8cm}p{1.8cm}p{1.8cm}} 
\hline\noalign{\smallskip}
 & EN &\multicolumn{2}{c} {AR} & \multicolumn{2}{c}{DE}   \\ 
  \noalign{\smallskip}\svhline\noalign{\smallskip}
  
 \textbf{model} & Ted (EN) & Alj.2h (AR) & Alj. (AR) & CV (DE) & Lect. (DE)\\ 
  \noalign{\smallskip}\svhline\noalign{\smallskip}

   % Mono-De  & 98,37   & 123,71 & 123,78 & \textbf{11,82} & \textbf{17,78}\\ 
   % Mono-Ar  & 100,62  & \textbf{9,74} & \textbf{16,00} & 103,22 & 104,35 \\
  %  Mono-En  & \textbf{7,58} & 167,52 & 160,3 & 140,24 & 112,07\\
    Mono-De  & -   & - & - & \textbf{11,82} & \textbf{17,78}\\ 
    Mono-Ar  & -  & \textbf{9,74} & \textbf{16,00} & - & - \\
    Mono-En  & \textbf{7,58} & - & - & - & -\\
    Mult. & 9,25 & 10,48 & 16,64 & 17,15 & 21,27 \\
\hline
\end{tabular}
\end{table}

\setlength{\tabcolsep}{2.4pt}
%\begin{table}[htbp]
\begin{table}
\caption{Results of baseline models on multilingual CS test sets. Results are reported in WER\%.}
\label{tab:res-bas-cs} 
\centering
\footnotesize
\begin{tabular}{p{1.5cm}p{1.5cm}p{1.5cm}p{1.5cm}p{1.5cm}p{1.5cm}p{1.5cm}p{1.5cm}}
 & \multicolumn{4}{c}{DE-EN}  & DE-AR & DE-AR-EN\\ \hline
 & \multicolumn{2}{c}{intra-sent.} & inter-sent. & mix-CS  &mix-CS & inter-sent \\ \hline
{\textbf{model}} & Deng. & SWC-CS & tst-inter  & D-E-CS & D-A-CS & artificial\\ \hline

    Mono-De  & \textbf{18,99}  & \textbf{28,97} & 48,55 & 50,83 & 64,00  & 83,87\\ 
    Mono-Ar  & 101,03 & 110,89 & 100,32 & 101,36 & 70,04 & 73,85\\
    Mono-En  & 104,60 & 118,83 & 63,47 & 67,11  & 117,14 & 84,89\\
    Mult. & 20,45 & 31,19 & \textbf{23,02}& \textbf{28,63} & \textbf{60,74} & \textbf{39,88}\\
\hline
\end{tabular}
\end{table}

\subsection{ Data augmented Code-Switching }
\label{subsec:cs}
In our first experiment, we trained a multilingual model using the data augmentation described in Section \ref{subsec:data-augmentation}.
Directly training the model with 50\% artificially created CS data leads to a bit more unstable gradients.
We trained the model multiple times.
While the performances were not that different the number of epochs needed for training was very different 109 and 198 for Mult.-noc1 and Mult.-noc2 Table~\ref{tab:res-c2-mono} respectively.
We reason the unstable gradients to be present due to the difficult data, as well as the nature of the task. While the Arabic language is Phonetically and script-wise very different from German or English, the quality of the used audio can also increase the difficulty of the task.

As mentioned in \cite{bengio2009curriculum} we apply a curriculum learning and first train on monolingual data, which can act as a regularization.
For the second stage of the curriculum, we took the multilingual model from Section \ref{subsec:baseline} and used the epoch with the lowest perplexity as a pre-trained model.
The same training hyper-parameters are applied as in the first training of the model and all weights are updated.
This model is denoted as Mult.cur50 and was trained with 50\% CS augmented data. 
This model was trained in only 39 Epochs compared to 109 Epochs without curriculum learning which shows a significantly faster convergence. 
We also applied the second curriculum step with only 20\% CS augmented data to see the effect it has on the training (Mult.-cur20).
As we have more updates with a higher learning rate in the two-stage approach we also trained the initial multilingual model a second time without CS data (Mult.-noCS).

The results of monolingual tests are shown in Table~\ref{tab:res-c2-mono}. 
As Mult.-cur20 has a slightly better performance compared to Mult.-cur50, we will focus on the model which was trained with 20\% CS augmented data.
We can see that training the CS models with the two-step approach yield the best performances and even outperform the monolingual models in Table ~\ref{tab:res-bas-mono}.
The only exception is the German CV test, however, while the Mult.-noCS model has a relative decrease of 17,93\% WER, training with CS mitigates the drop in performance to only a 9,64\% decrease compared to Mono-DE.
\setlength{\tabcolsep}{2.4pt}
%\begin{table}[htbp]
\begin{table}
\caption{Results of multilingual models on monolingual test sets. Results are reported in WER\%.}
\label{tab:res-c2-mono} 
%\small
\begin{tabular}{p{1.7cm}p{1.8cm}p{1.8cm}p{1.8cm}p{1.8cm}p{1.8cm}p{1.8cm}} 
\hline\noalign{\smallskip}
 & EN &\multicolumn{2}{c} {AR} & \multicolumn{2}{c}{DE}   \\ 
  \noalign{\smallskip}\svhline\noalign{\smallskip}
 \textbf{model} & Ted (EN) & Alj.2h (AR) & Alj. (AR) & CV (DE) & Lect. (DE)\\ 
  \noalign{\smallskip}\svhline\noalign{\smallskip}
    Mult.          & 9,25              & 10,48         & 16,64             & 17,15             & 21,27             \\
    Mult.-noCS     & 7,76              & 10,22         & 15,44             & 13,94             & 17,84             \\
    Mult.-noc1     & 7,77              & 9,84          & 15,82             & 14,93             & 18,68             \\ 
    Mult.-noc2     & 8,67              & 10,70         & 16,82             & 16,76             & 20,87             \\
    Mult.-cur50    & \textbf{7,12}     & 9,32          & \textbf{15,11}    & 13,23             & 17,82             \\
    Mult.-cur20    & 7,14              & \textbf{9,30} & 15,33             & \textbf{12,96}    & \textbf{17,32}    \\
\hline
\end{tabular}
\end{table}

\begin{figure}[b]
%\sidecaption
\includegraphics[scale=0.20,page=3,trim={6cm 7cm 8cm 7cm},clip]{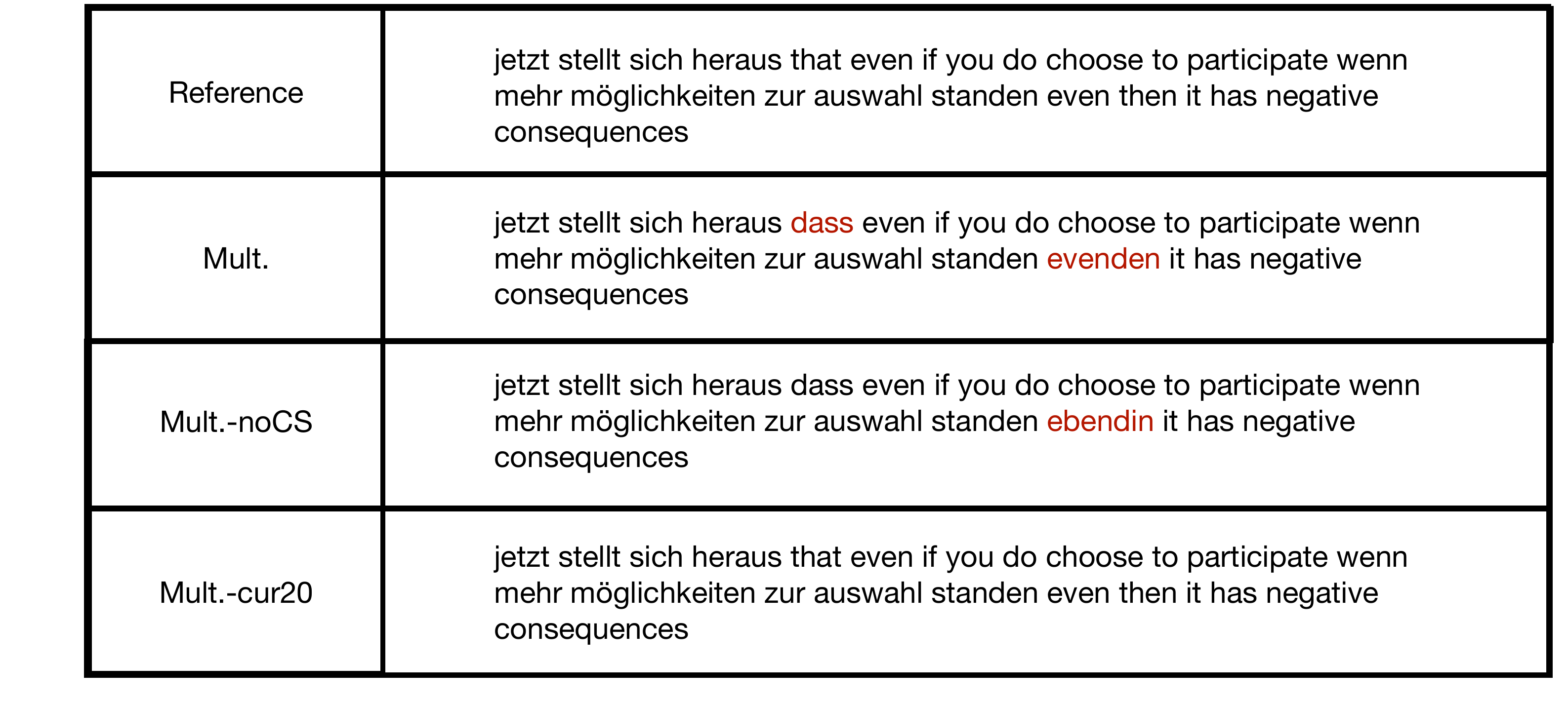}
\caption{Transcription hypothesis for the Referenz: "jetzt stellt sich heraus that even if you do choose to participate wenn mehr möglichkeiten zur auswahl standen even then it has negative consequences" German parts are (now it turns out) (when there are more choices present)}
\label{fig:examples}       % Give a unique label
\end{figure}

In Table~\ref{tab:res-c2-multi} the CS results after the second-curriculum are depicted.
On our own small intra-sentential Denglish set we see that training without curriculum (Mult.-noc1) hurts the performance, compared to the Mult.-noCS model which was trained without data augmentation.
The other data-augmented models can roughly keep the same WER.
On the bigger German-English intra-sentential SWC-CS test set we can observe a relative improvement of 2,27\% and 2,58\% WER for the Mult.-cur20 and Mult.-cur50 models over the baseline multilingual model (Mult.-noCS).
%On the bigger German-English intra-sentential SWC-CS test set we can observe a relative improvement of 2.27\% and 2,58\% WER for the Mult.-cur20 and Mult.-cur50 models over the two-step trained baseline multilingual model.
More importantly, however, compared to training without CS data, utilizing a CS augmentation of 20\% yields relative improvements of 10,76\% WER on the tst-inter data and a relative improvement of 8,55\% WER on the D-E-CS test set, as well as a relative improvement of 25,26\% on D-A-CS.
Similar to previous work we also evaluated our models on artificially created CS data with switches between all languages (DE-AR-EN). 
The Mult.-cur20 yields a relative improvement of 80,35\% WER compared to the multilingual model without CS (Mult.-noCS), which is extremely high when compared to our collected in-house test data.
This is why we ignore this test case in our ablation studies, as we believe that testing on artificial data yields inflating improvements, which do not hold on our collected data, although it is only read speech.
\setlength{\tabcolsep}{2.4pt}
%\begin{table}[htbp]
\begin{table}
\caption{Results of multilingual models on CS test sets. Results are reported in WER\%.}
\label{tab:res-c2-multi} 
%\scriptsize
\footnotesize
\begin{tabular}{p{1.5cm}p{1.5cm}p{1.5cm}p{1.5cm}p{1.5cm}p{1.5cm}p{1.5cm}p{1.5cm}}
 & \multicolumn{4}{c}{DE-EN}  & DE-AR & DE-AR-EN\\ \hline
 & \multicolumn{2}{c}{intra-sent.} & inter-sent. & mix-CS  &mix-CS & inter-sent \\ \hline
{\textbf{model}} & Deng. & SWC-CS & tst-inter  & D-E-CS & D-A-CS & artificial\\
  \noalign{\smallskip}\svhline\noalign{\smallskip}

    Mult.          & 20,45             & 31,19             & 23,02             & 28,63             & 60,74         & 39,88\\  
    Mult.-noCS     & 16,38             & 28,64             & 20,91             & 25,98             & 53,90         & 44,32\\
    Mult.-noc1     & 18,28             & 28,98             & 20,12             & 25,47             & 55,57         & 9,25\\ 
    Mult.-noc2     & 19,30             & 31,24             & 19,82             & 26,79             & 54,97         & 10,73\\
    Mult.-cur50    & 1\textbf{6,23}    & 27,99             & 18,81             & \textbf{23,63}    & 45,67         & \textbf{8,66}\\
    Mult.-cur20    & 16,40             & \textbf{27,90}    & \textbf{18,66}    & 23,76             & \textbf{45,40}& 8,71\\
\hline
\end{tabular}
\end{table}
Fig.~\ref{fig:examples} shows an example output on the tst-inter test set.
%\begin{eqnarray*}
%&\textrm{Reference: "that did the british for example not at all during the german blitz}\\ &\textrm{dort gab s entweder behelfsmäßige"}\\
%&\textrm{Mult.: "that did the british for example not at arguing the german blitz}\\
%&\textrm{dot gaps entweder the health message"}\\
%&\textrm{Mult.-noCS: "that did the british for example not at all during the german blitz}\\
%&\textrm{dotgups entweder perhaps missing it"}\\
%&\textrm{Mult.-cur20: "that did the british for example not at all during the german blitz}\\
%&\textrm{dot gabs entweder behelfsmäßige"} \\
%\end{eqnarray*}
%\begin{eqnarray*}
%&\textrm{Ref.: "jetzt stellt sich heraus that even if you do choose to participate}\\ 
%&\textrm{wenn mehr möglichkeiten zur auswahl standen even then it has negative consequences"}\\
%&\textrm{Mult.: "jetzt stellt sich heraus \textit{dass} even if you do choose to participate}\\
%&\textrm{wenn mehr möglichkeiten zur auswahl standen \textit{evenden} it has negative consequences"}\\
%&\textrm{Mult.-noCS: "jetzt stellt sich heraus dass even if you do choose to participate}\\
%&\textrm{wenn mehr möglichkeiten zur auswahl standen \textit{ebendin} it has negative consequences"}\\
%&\textrm{Mult.-cur20: "jetzt stellt sich heraus that even if you do choose to participate}\\
%&\textrm{wenn mehr möglichkeiten zur auswahl standen even then it has negative consequences"}\\
%\end{eqnarray*}
This example shows that the model is now more reliable when it comes to switching the language when transcribing at the switching region.
The difficulty of such transcription may also arise due to people speaking English with an accent of their first language.
In general, the model trained with pseudo-CS data seems more reliable when transcribing words at switching points in the utterance.

The results depicted in Table~\ref{tab:res-c2-mono} and Table~\ref{tab:res-c2-multi} show that using CS augmented data does not just improve models on CS data but also improves the model's performance over the monolingual model on the respective monolingual test sets, as well.

\subsection{Ablation studies}
\label{subsubsec:zero-shot-cs}
After seeing the results in Section~\ref{subsec:cs} we further analyzed the effect of utilizing this kind of artificially created pseudo-CS data during training. 
Specifically, in a scenario with many more languages, the question will arise if we need to ensure generating CS data with transitions between all possible languages, and do we need to ensure that bidirectional transitions from one language to all others need to be present to enable CS during inference? 

For this reason, we conducted several further experiments.
All experiments apply the same curriculum learning regime and use the multilingual model described in Section~\ref{subsec:baseline} as a starting point. 
The results are depicted in Table~\ref{tab:res-c2-mono-zero} and Table~\ref{tab:res-c2-multi-zero}. 
The ending of model names depicts which transitions were not seen during training, for example, "nodear" means there was no transition from German to Arabic. 
"nodex" means that German utterances were not used in any CS data.
In contrast, "odex" depicts the case, in which only transitions from and to German were present. %and there was no switch between English and Arabic.
%\todo{Revise: Everything after this needs to be revised according to new evaluations}

The results of monolingual tests Table~\ref{tab:res-c2-mono-zero} give interesting insights into using artificially created CS data for multilingual models. 
We can see that all models which saw CS data during training outperform the baseline multilingual model (Mult.-noCS). 
We can appreciate that usually depending on which language or language transition was kept out of the training the performance on the respective test seems to degrade slightly compared to the Mult.-cur20 model. 
The reason is that these restrictions make the other languages proportionally more present in the training data. 
This is also supported by the WER improvements on the other languages which were not restricted. 
An example would be the performance of Mult.-nodeen (third row) on the TED performance and the Alj.2h set.
Compared to Mult.-cur20 (second row) the WER on Ted decreased from 7,14\% WER to 7,21\%, while the performance on Alj.2h improved from 9,30\% WER to 9,12\%.
\setlength{\tabcolsep}{2.4pt}
%\begin{table}[htbp]
\begin{table}
\caption{Results of multilingual models on monolingual test sets. Models were trained using 20\% data augmentation with varying restrictions. Results are reported in WER\%.}
\label{tab:res-c2-mono-zero}
%\small
\begin{tabular}{p{1.7cm}p{1.8cm}p{1.8cm}p{1.8cm}p{1.8cm}p{1.8cm}p{1.8cm}} 
\hline\noalign{\smallskip}
 & EN &\multicolumn{2}{c} {AR} & \multicolumn{2}{c}{DE}   \\ 
  \noalign{\smallskip}\svhline\noalign{\smallskip}
 \textbf{model} & Ted (EN) & Alj.2h (AR) & Alj. (AR) & CV (DE) & Lect. (DE)\\ 
  \noalign{\smallskip}\svhline\noalign{\smallskip}
    %Mult. &  9,25 & 7,62 & 10,48 & 16,64 & 17,15 & 21,27 \\
    Mult.-noCS & 7,76  & 10,22 & 15,44 & 13,94 & 17,84 \\
    Mult.-cur20  & \textbf{7,14} & 9,30 & 15,33 & 12,96 & 17,32 \\
    \hline
    Mult.-nodeen   & 7,21          & \textbf{9,12} & 15,08             & 13,08         & 17,68     \\ 
    Mult.-nodear   & 7,19          & 9,23          & 15,15             & 12,84         & 17,36     \\ 
    Mult.-nodex    & 7,23          & 9,26          & 15,17             & 13,39         & 17,80     \\
    Mult.-odex     & 7,20          & 9,43          & \textbf{14,90}    & 12,88         & 17,20     \\ 
    Mult.-noende   & 7,28          & 9,13          & 15,12             & 13,11         & 17,42     \\ 
    Mult.-noenar   & 7,22          & 9,14          & 15,06             & 12,86         & 16,99     \\ 
    Mult.-noenx    & 7,32          & 9,36          & 15,06             & 12,86         & 17,46     \\ 
    Mult.-oenx     & 7,28          & 9,42          & 15,34             & 12,91         & 17,42     \\ 
    Mult.-noarde   & 7,29          & 9,14          & 15,00             & 12,98         & 17,58     \\ 
    Mult.-noaren   & 7,20          & 9,44          & 15,52             & 12,89         & 17,31     \\ 
    Mult.-noarx    & 7,17          & 9,57          & 15,18             & \textbf{12,68}& \textbf{16,92}\\ 
    Mult.-oarx     & 7,19          & 9,01          & 15,06             & 12,97         & 17,00     \\ 
\hline
\end{tabular}
\end{table}

Looking at intra-sent. evaluations, specifically the Deng. test, Table ~\ref{tab:res-c2-multi-zero},  we can observe similar behaviour to Table~\ref{tab:res-c2-mono-zero}.
%Looking at Table ~\ref{tab:res-c2-multi-zero} we can see similar behaviour to Table~\ref{tab:res-c2-mono-zero} when it comes to intra-sent. sets.
In the second last row x-noarx the model trained with only German and English switches is depicted. 
Compared to x-cur20 (second row), this results in more German and English utterances being seen during training and thus slightly improves over the model by 3,48\% relative WER.
This is due to the intra-sent. test set only containing examples with German as the matrix language and English words embedded.

Another interesting point for Intra-sentential CS is the phenomenon, in which words from the embedded language, in our case English, can happen to be adapted according to the grammar of the matrix language.
This being the case we also report the accuracy of correctly transcribed English words in our Deng. test set and report them in the Column (Deng.Acc) in Table~\ref{tab:res-c2-multi-zero}.
From these numbers, we can see that there is an inverse correlation between correctly transcribing English words and the overall WER on Deng. test data.
However, as the data augmentation only uses CS between longer clauses we see that there is only a limited effect on such intra-sentential data.

When looking at inter-sent. and mix-CS examples for DE-EN language pairs, we can see that the model trained without German switches, x-nodex (fifth row), decreases performance compared to x-cur20.
The performance on tst-inter decreased from 18,66\% WER to 20,38\%.
However, it still performs slightly better than the baseline multilingual (x-noCS) model with 20,91\% WER.
Models which never saw switches from German to English, x-nodeen (third row), also lose a bit of performance compared to x-cs20.
However, looking at the transcriptions we can see that the model is still able to transcribe switches from German to English, which shows that the model is able to generalize the possibility of switching between languages and not just learns one specific switch.
The answer to one of the questions of this ablation study is very well answered on the D-A-CS test data.
Here the worst performing model which has seen any kind of Arabic CS data during training is the x-oenx model (tenth row) which only saw switches from and to English. 
On the D-A-CS test which only contained Arabic and German CS data, the performance of the model improves over the baseline multilingual model relative by 9,33\% WER although the model never saw switches between aforementioned languages.
This shows that when training models to transcribe CS, especially in the inter-sentential case there is no need to provide switches between all language pairs.

% Intra-sentential CS is a very interesting phenomenon, in which words from the embedded language, in our case English, can happen to be adapted according to the grammar of the matrix language.
% This being the case we also report the accuracy of correctly transcribed English words in our Denglish test set and report them in the Column (Deng.Acc) in Table~\ref{tab:res-c2-multi-zero}.
% From these numbers, we can see that there is an inverse correlation between correctly transcribing English words and the overall WER on Deng. test data.
% However, as the data augmentation only uses CS between longer clauses we see that there is only a limited effect on such intra-sentential data.

Considering the mentioned results, we can appreciate that the model generally benefits from seeing CS data during training. 
It not just improves monolingual performance but also improves on inter-sentential CS data.
We can also see that the model has the general capability to learn to switch between languages never seen in a CS scenario during training.
However, at least seeing the language switched one time with any other language greatly improves over not switching at all. 
Using each language at least once in any switching combination can massively improve the capability of the model in general, no matter if a specific language switch was seen during training or not.

\setlength{\tabcolsep}{2.4pt}
%\begio{table}[htbp]
\begin{table}
\caption{Results of multilingual models on CS test sets. Models were trained using 20\% data augmentation with varying restrictions. Results are reported in WER\%.  Deng.Acc is the accuracy of correctly transcribed English words in percentage.}
\label{tab:res-c2-multi-zero} 
\small
\begin{tabular}{p{1.5cm}p{1.5cm}p{1.5cm}p{1.5cm}p{1.5cm}p{1.5cm}p{1.5cm}}
 % & \multicolumn{4}{c}{DE-EN}  & DE-AR & DE-AR-EN\\ \hline
 % & \multicolumn{2}{c}{intra-sent.} &  &inter-sent.   &mix-CS & inter-sent \\ \hline
  & \multicolumn{5}{c}{DE-EN}  & DE-AR \\ \hline
 & \multicolumn{2}{c}{intra-sent.} &  &inter-sent.   &mix-CS & mix-CS \\ \hline
{\textbf{Mult.}} & Deng. & Deng.Acc & SWC-CS & tst-inter & D-E-CS & D-A-CS\\ \hline
   % Mult. & 21,83 & 31,19 & 22,42 & 39,88 \\
    x-noCS     & 16,38   & 79,03          & 28,64             & 20,91             & 25,98             & 53,90     \\
    x-cur20    & 16,40   & 79,53          & 27,90             & 18,66             & 23,76             & 45,40     \\
    \hline                          
    x-nodeen   & 16,23   & 79,87          & 28,00             & 19,36             & 25,02             & 43,81     \\
    x-nodear   & 16,34   & 80,87          & 27,92             & 18,19             & 23,30             & 46,85     \\
    x-nodex    & 16,95   & 79,19          & 28,47             & 20,38             & 25,63             & 51,26     \\
    x-odex     & 16,06   & 81,20          & 27,89             & \textbf{17,39}    & 23,92             & 44,36     \\
    x-noende   & 17,30   & 79,03          & 27,99             & 18,58             & 24,33             & 43,18     \\
    x-noenar   & 16,18   & 80,03          & \textbf{27,71}    & 18,61             & 24,55             & 46,22     \\
    x-noenx    & 16,19   & 79,87          & 27,98             & 21,58             & 27,29             & 41,84     \\
    x-oenx     & 16,39   & 80,87          & 27,99             & 18,21             & 23,85             & 48,87     \\
    x-noarde   & 16,38   & 80,20          & 28,09             & 18,08             & \textbf{23,21}    & 46,96     \\
    x-noaren   & 16,23   & 80,54          & 27,95             & 18,53             & 24,10             & 45,04       \\
    x-noarx    & \textbf{15,83} & \textbf{81,71}   & 27,80             & 18,00             & 24,43             & 55,77     \\
    x-oarx     & 16,54    & 80,37         & 28,10             & 19,66             & 25,25             & \textbf{41,05}    \\
\hline
\end{tabular}
\end{table}

\subsubsection{Transformer architecture}
\label{subsubsec:transformer}
In order to see the if this data-augmentation is generalisable we trained a Transformer based S2S model, as well.
%Another interesting aspect of this work would be to know if the augmentation strategy can be applied to other architectures as well. 
%Thus we also trained a Transformer based S2S model.
The model consists of two CNN layers and six encoder and four decoder layers with a hidden size of 1024.
%\setlength{\tabcolsep}{2.4pt}
%%\begin{table}[htbp]
%\begin{table}
%\caption{Results of multilingual Transformer models on monolingual test sets. Results are reported in WER\%.}
%\label{tab:Trans-c2-mono} 
%%\small
%\begin{tabular}{p{1.7cm}p{1.8cm}p{1.8cm}p{1.8cm}p{1.8cm}p{1.8cm}p{1.8cm}} 
%\hline\noalign{\smallskip}
% & EN &\multicolumn{2}{c} {AR} & \multicolumn{2}{c}{DE}   \\ 
%  \noalign{\smallskip}\svhline\noalign{\smallskip}
% \textbf{model} & Ted (EN) & Alj.2h (AR) & Alj. (AR) & CV (DE) & Lect. (DE)\\ 
%  \noalign{\smallskip}\svhline\noalign{\smallskip}
%    T.Mult.          & 10,50              & 13,07         & 19,44            & 18,74             & 22,21             \\
%    T.Mult.-noCS     & 10,13              & 12,62         & 18,93             & 17,93             & 21,27             \\
%    T.Mult.-cur20    & \textbf{9,81}      & \textbf{12,03} & \textbf{17,90}  & \textbf{17,33}    & \textbf{20,94}    \\
%\hline
%\end{tabular}
%\end{table}

\setlength{\tabcolsep}{2.4pt}
%\begin{table}[htbp]
\begin{table}
\caption{Results of multilingual Transformer models on CS test sets. Results are reported in WER\%.}
\label{tab:Trans-c2-multi} 
%\scriptsize
\footnotesize
\begin{tabular}{p{1.7cm}p{1.8cm}p{1.8cm}p{1.8cm}p{1.8cm}p{1.8cm}p{1.8cm}}
 % & \multicolumn{4}{c}{DE-EN}  & DE-AR & DE-AR-EN\\ \hline
 % & \multicolumn{2}{c}{intra-sent.} &  &inter-sent.   &mix-CS & inter-sent \\ \hline
  & \multicolumn{4}{c}{DE-EN}  & DE-AR \\ \hline
 & intra-sent. &  &inter-sent.   &mix-CS & mix-CS \\ \hline
{\textbf{Mult.}} & Deng. & SWC-CS & tst-inter & D-E-CS & D-A-CS\\ \hline
  \noalign{\smallskip}\svhline\noalign{\smallskip}

    T.Mult.          & 20,87             & 33,33             & 24,20             & 32,27             & 54,59         \\  
    T.Mult.-noCS     & 20,93             & 32,67             & 23,54             & 31,03             & 53,61         \\
    T.Mult.-cur20    & \textbf{20,55}    & \textbf{31,92}    & \textbf{20,01}    & \textbf{26,70}    & \textbf{53,06}\\
\hline
\end{tabular}
\end{table}
%In Table~\ref{tab:Trans-c2-mono}
Due to restricted space we only display the multilingual results, however, the monolingual results are similar to the LSTM-based model, as well.
In Table~\ref{tab:Trans-c2-multi} it is possible to see that the general trends from the LSTM-based model also hold for the Transformer.
The model performs worse than our LSTM architecture, however, this might be due to suboptimal hyperparameters as we focused on the LSTM model for our experiments.

\section{Conclusion}
\label{sec:conclusion}
In this work, we described a simple yet effective way of artificially generating CS data to improve on the inter-sentential CS task.
We showed that our collected read-speech test data is more reliable for performance evaluation than using artificially generated test data.
We also saw that the presented approach improves the monolingual performance of multilingual models, without any changes in the model architecture.
More importantly, we enable a language-agnostic multilingual S2S model to automatically transcribe CS speech without providing any real CS data. 
Our experiments reveal that a model trained on artificial pseudo-CS data between language $x\leftrightarrow  y$ and $y\leftrightarrow z$ is able to transcribe CS utterances with switches between languages $x\leftrightarrow z$.
In such a scenario our model x-oenx (tenth row) Table~\ref{tab:res-c2-multi-zero} improves over the baseline multilingual model x-noCS by 5,03\%WER, column D-A-CS.
These results are especially important as there are millions of multilingual speakers code-switching in their everyday life.
Thus systems able to process these inputs are much needed even when there is no data for all language pairs.
In the future, we want to use language pairs from previous work in order to be able to compare to those as well.
\section{Acknowledgement}
The project on which this report is based was funded by the Federal Ministry of Education and Research (BMBF) of Germany under the numbers 01EF1803B (RELATER) and 01IS18040A (OML).

%\bibliographystyle{thebibliography}
%\bibliography{templates/refslb}

%\input{templates/refsl}

%\input{referenc}

%

\backmatter%%%%%%%%%%%%%%%%%%%%%%%%%%%%%%%%%%%%%%%%%%%%%%%%%%%%%%%
\appendix
%%%%%%%%%%%%%%%%%%%%% appendix.tex %%%%%%%%%%%%%%%%%%%%%%%%%%%%%%%%%
%
% sample appendix
%
% Use this file as a template for your own input.
%
%%%%%%%%%%%%%%%%%%%%%%%% Springer-Verlag %%%%%%%%%%%%%%%%%%%%%%%%%%

\chapter{Chapter Heading}
\label{introA} % Always give a unique label
% use \chaptermark{}
% to alter or adjust the chapter heading in the running head

Use the template \emph{appendix.tex} together with the Springer document class SVMono (monograph-type books) or SVMult (edited books) to style appendix of your book in the Springer layout.

\section{Section Heading}
\label{sec:A1}
% Always give a unique label
% and use \ref{<label>} for cross-references
% and \cite{<label>} for bibliographic references
% use \sectionmark{}
% to alter or adjust the section heading in the running head
Instead of simply listing headings of different levels we recommend to let every heading be followed by at least a short passage of text. Further on please use the \LaTeX\ automatism for all your cross-references and citations.

\subsection{Subsection Heading}
\label{sec:A2}
Instead of simply listing headings of different levels we recommend to let every heading be followed by at least a short passage of text. Further on please use the \LaTeX\ automatism for all your cross-references and citations as has already been described in Sect.~\ref{sec:A1}.

For multiline equations we recommend to use the \verb|eqnarray| environment.
\begin{eqnarray}
\vec{a}\times\vec{b}=\vec{c} \nonumber\\
\vec{a}\times\vec{b}=\vec{c}
\label{eq:A01}
\end{eqnarray}

\subsubsection{Subsubsection Heading}
Instead of simply listing headings of different levels we recommend to let every heading be followed by at least a short passage of text. Further on please use the \LaTeX\ automatism for all your cross-references and citations as has already been described in Sect.~\ref{sec:A2}.

Please note that the first line of text that follows a heading is not indented, whereas the first lines of all subsequent paragraphs are.

\begin{figure}[t]
\sidecaption[t]
% For example, with the graphicx style use
\includegraphics[scale=.65]{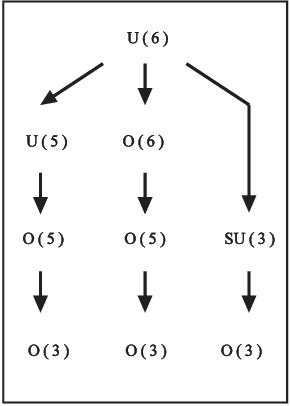}
%
% If no graphics program available, insert a blank space i.e. use
%\picplace{5cm}{2cm} % Give the correct figure height and width in cm
%
\caption{Please write your figure caption here}
\label{fig:A1}       % Give a unique label
\end{figure}

\begin{table}
\caption{Please write your table caption here}
\label{tab:A1}       % Give a unique label
%
% Follow this input for your own table layout
%
\begin{tabular}{p{2cm}p{2.4cm}p{2cm}p{4.9cm}}
\hline\noalign{\smallskip}
Classes & Subclass & Length & Action Mechanism  \\
\noalign{\smallskip}\hline\noalign{\smallskip}
Translation & mRNA$^a$  & 22 (19--25) & Translation repression, mRNA cleavage\\
Translation & mRNA cleavage & 21 & mRNA cleavage\\
Translation & mRNA  & 21--22 & mRNA cleavage\\
Translation & mRNA  & 24--26 & Histone and DNA Modification\\
\noalign{\smallskip}\hline\noalign{\smallskip}
\end{tabular}
$^a$ Table foot note (with superscript)
\end{table}
%

%%%%%%%%%%%%%%%%%%%%%%acronym.tex%%%%%%%%%%%%%%%%%%%%%%%%%%%%%%%%%%%%%%%%%
% sample list of acronyms
%
% Use this file as a template for your own input.
%
%%%%%%%%%%%%%%%%%%%%%%%% Springer %%%%%%%%%%%%%%%%%%%%%%%%%%

\Extrachap{Glossary}

Use the template \emph{glossary.tex} together with the Springer document class SVMono (monograph-type books) or SVMult (edited books) to style your glossary\index{glossary} in the Springer layout.

\runinhead{glossary term} Write here the description of the glossary term. Write here the description of the glossary term. Write here the description of the glossary term.

\runinhead{glossary term} Write here the description of the glossary term. Write here the description of the glossary term. Write here the description of the glossary term.

\runinhead{glossary term} Write here the description of the glossary term. Write here the description of the glossary term. Write here the description of the glossary term.

\runinhead{glossary term} Write here the description of the glossary term. Write here the description of the glossary term. Write here the description of the glossary term.

\runinhead{glossary term} Write here the description of the glossary term. Write here the description of the glossary term. Write here the description of the glossary term.
\printindex

%%%%%%%%%%%%%%%%%%%%%%%%%%%%%%%%%%%%%%%%%%%%%%%%%%%%%%%%%%%%%%%%%%%%%%

\end{document}